\documentclass{article}

\usepackage[preprint]{neurips_2024}

\usepackage[utf8]{inputenc} 
\usepackage[T1]{fontenc}    
\usepackage{hyperref}       
\usepackage{url}            
\usepackage{booktabs}       
\usepackage{amsmath,amsfonts,amssymb,amsthm}       
\usepackage{nicefrac}       
\usepackage{microtype}      
\usepackage{xcolor}         
\usepackage{xspace}
\usepackage{graphicx}
\usepackage{booktabs}
\usepackage{multirow}
\usepackage[inline]{enumitem}
\usepackage{subcaption}

\newcommand{\myclip}{AlignCLIP\xspace}
\newcommand{\sharedclip}{SharedCLIP\xspace}
\newcommand{\imsep}{IMSep\xspace}

\newcommand{\trainingdata}{CC12M\xspace}

\newcommand{\norm}[1]{\left\lVert#1\right\rVert}

\newcommand{\evi}{\vec{e}_\mathrm{v}^{\,i}}
\newcommand{\eti}{\vec{e}_\mathrm{t}^{\,i}}
\newcommand{\evj}{\vec{e}_\mathrm{v}^{\,j}}
\newcommand{\etj}{\vec{e}_\mathrm{t}^{\,j}}

\newcommand{\esi}{\vec{e}_\mathrm{s}^{\,i}}

\newcommand{\Ev}{E_\mathrm{v}}
\newcommand{\Et}{E_\mathrm{t}}
\newcommand{\Es}{E_\mathrm{s}}
\newcommand{\Evt}{\mathcal{M}}
\newcommand{\Evv}{\mathcal{V}}

\newcommand{\Evs}{\mathcal{V_D}}

\newcommand{\yv}{\hat{y}_\mathrm{v}}
\newcommand{\yt}{\hat{y}_\mathrm{t}}

\newcommand{\yvsep}{\hat{y}_\mathrm{vsep}}

\newcommand{\transpose}{\intercal}
\setcitestyle{round}
\let\cite\undefined

\title{Mitigate the Gap: Investigating Approaches for Improving Cross-Modal Alignment in CLIP}

\author{%
  Sedigheh Eslami\\
  Hasso Plattner Institute\\
 \texttt{sedigheh.eslami@hpi.de} \\
  \And
  Gerard de Melo \\
  Hasso Plattner Institute\\
  \texttt{gerard.demelo@hpi.de} \\
}

\begin{document}

\maketitle

\begin{abstract}
  Contrastive Language--Image Pre-training (CLIP) has manifested remarkable improvements in zero-shot classification and cross-modal vision-language tasks. Yet, from a geometrical point of view, the CLIP embedding space has been found to have a pronounced modality gap. This gap renders the embedding space overly sparse and disconnected, with different modalities being densely distributed in distinct subregions of the hypersphere. In this work, we aim at answering three main questions: 1.\ Does sharing the parameter space between the multi-modal encoders reduce the modality gap? 2.\ Can the gap be mitigated by pushing apart the uni-modal embeddings via intra-modality separation? 3.\ How do these gap reduction approaches affect the downstream performance? We design \myclip, in order to answer these questions and through extensive experiments, we show that \myclip achieves noticeable enhancements in the cross-modal alignment of the embeddings, and thereby, reduces the modality gap, while improving the performance across several zero-shot and fine-tuning downstream evaluations. The source code for reproducing our experiments is available at \href{https://github.com/sarahESL/AlignCLIP}{https://github.com/sarahESL/AlignCLIP}.
\end{abstract}

\section{Introduction} 
 One of the most prominent and widely used pre-trained vision--language models is OpenAI's Contrastive Language--Image Pre-training (CLIP) model \citep{clip}. CLIP is a dual-stream vision--language encoder trained for learning a shared representation space, in which image and text modalities can be jointly embedded.
 It has demonstrated exceptional zero-shot capabilities for image classification, multi-modal retrieval as well as robustness to natural distribution shifts.
 
\begin{figure}[t]
\centering
  \includegraphics[width=0.98\linewidth]{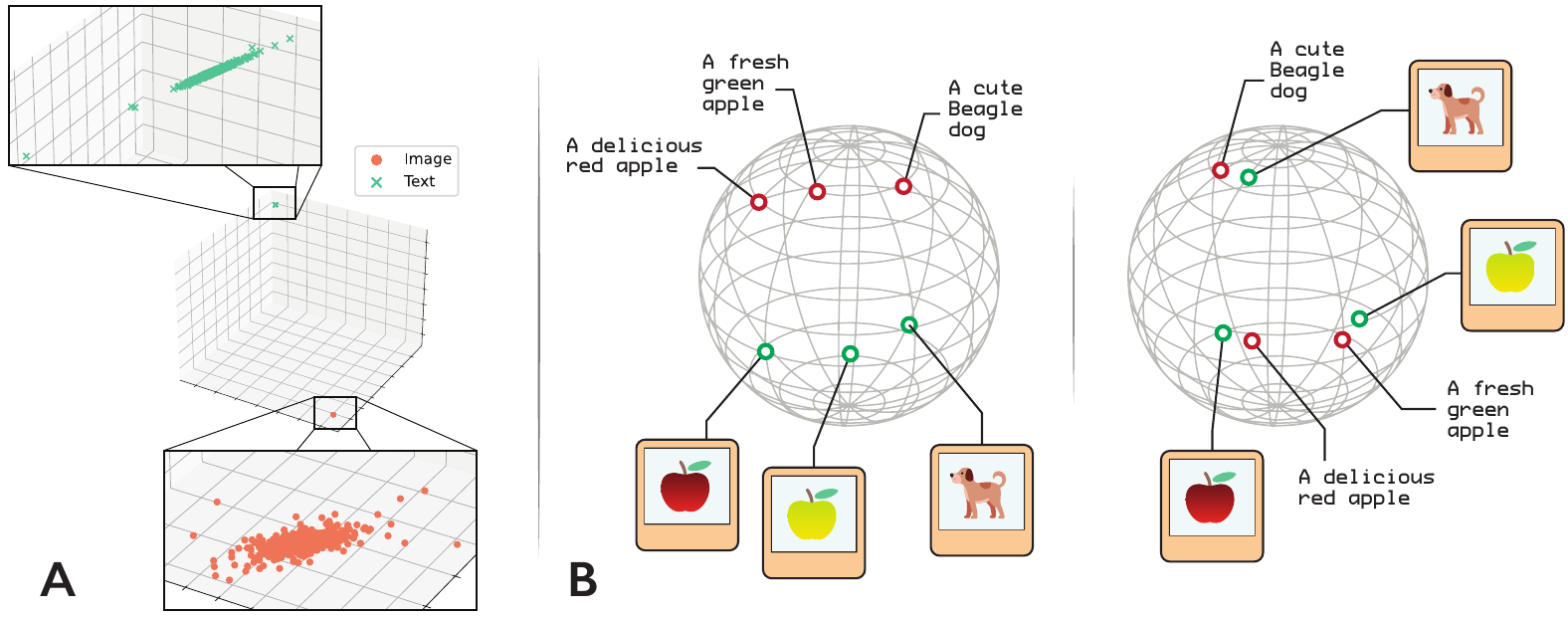}
  \caption{Modality gap and cross-modal alignment. (A) The average pairwise cosine similarity of the encoded image--text pairs is about 0.22, i.e., the average angle
  is about 78 degrees. (B) Schematic illustration of unaligned (left) versus aligned (right) embedding spaces.}
  \label{fig-intro-alignment}
\end{figure}
 Despite the outstanding performance of CLIP, recent work has shed light on a pronounced \emph{modality gap} in the CLIP embedding space \citep{mind_gap, isotropy, two_effects}, leading to large distances between image and text embeddings. We illustrate these phenomena more comprehensively in Figure~\ref{fig-intro-alignment}A, in which the DOSNES \citep{dosnes} projection of the CLIP-encoded image--text pairs from CC3M \citep{cc3m} using the pre-trained ViT-B-32 backend is shown. As can be seen, each modality densely populates a separate small subregion of the CLIP's representation embedding space. On the contrary, with a meaningful and well-structured cross-modal embedding space, we envision having similar samples fairly closely aligned with each other, regardless of their modalities. 
  An example of such an embedding space is shown in Figure~\ref{fig-intro-alignment}B, on the right side, with similar images and texts closely located on the hypersphere. In contrast, the left side of Figure~\ref{fig-intro-alignment}B shows an example of an unaligned embedding space, inspired by the CLIP visualization.
 
 Through extensive experiments, \citet{mind_gap} and \citet{isotropy} showed that the modality gap is caused by a combination of the model initialization and the contrastive loss optimization.   
 Recently, \citet{two_effects} showed that the driving factor behind the modality gap is the information-imbalance between the two modalities, i.e., texts in the training datasets usually have less detailed information in comparison to their pairing images.
 
 Attempting to reduce the modality gap, or equivalently, increase the cross-modal alignment in CLIP while enhancing the performance in downstream tasks, under the same amount of information-imbalance, we propose \myclip. Ultimately, we answer three main questions:  
\begin{enumerate*}
    \item With the same training dataset, i.e., under the same amount of information-imbalance between the two modalities, does sharing the parameter space between the multi-modal encoders reduce the modality gap?
    \item Can the gap be further mitigated by pushing apart the uni-modal embeddings via intra-modality separation?
    \item How do these gap reduction approaches affect the downstream performance?
\end{enumerate*}
 
  We answer the first question by sharing the transformer encoder and the projection layer in the vision and language encoders, and observe that it already results in noticeable cross-modal alignment improvements as well as performance increase. As for the second question, we introduce the Intra-Modality Separation objective that encourages the embeddings within the visual modality to be expanded and pushed towards the language modality. As a result, each modality gets moved towards the embeddings from the opposite modality and thereby, the modality gap can get reduced. Our experimental results support that the two aforementioned refinements show substantial improvement of the cross-modal alignment while improving the zero-shot transfer performance across a wide variety of downstream tasks as well as linear probing for image classification and fine-tuning in cross-modal retrieval. As apposed to previous work that attempted at reducing the gap with naive isomorphic translations with respect to the distance of image-text pairs \citep{mind_gap, two_effects, isotropy}, which can hurt the distances of the unpaired samples, and therefore, distort the meaningful structure of the embedding space, \myclip reduces the gap by modifications that improve the semantic structure of the latent modality space, as motivated by \citep{understanding_structure}.

\section{Background, Notations and Concepts}
Given a set of $N$ image--text pairs, 
we consider the CLIP image encoder to obtain the $l_2$-normalized vector $\evi \in \mathbb{R}^d$ as a $d$-dimensional embedding vector for image $v_i$ and the CLIP text encoder to obtain the $l_2$-normalized text embedding $\eti \in \mathbb{R}^d$ for the text sample $t_i$. We denote a batch of encoded image--text pairs by $\Ev \in \mathbb{R}^{b \times d}$ for images and $\Et \in \mathbb{R}^{b \times d}$ for texts, where $b$ refers to the batch size.

\vspace{1mm}
\noindent \textbf{Background on CLIP}. The contrastive objective in CLIP is the average of the vision to language and the language to vision Info-NCE contrastive loss functions 
formulated as:
\begin{equation}
\label{eq-clip-vl}
   \mathcal{L}_{v \rightarrow l} = \frac{-1}{N} \sum_{i=1}^{N} \log \frac{\exp[(\evi \cdot \eti)\mathbin{/}\tau]}{\sum_{j=1}^N \exp[(\evi \cdot \etj)\mathbin{/}\tau]}, \quad \mathcal{L}_{l \rightarrow v} = \frac{-1}{N} \sum_{j=1}^{N} \log \frac{\exp[(\evj \cdot \etj)\mathbin{/}\tau]}{\sum_{i=1}^N \exp[(\evi \cdot \etj)\mathbin{/}\tau]},
\end{equation}
respectively, where $\tau$ is the learnable temperature parameter. 
The overall CLIP loss is then:
\begin{equation}
\label{eq-clip-total}
    \mathcal{L}_{\textrm{clip}} = \frac{1}{2} \left[\mathcal{L}_{v \rightarrow l} + \mathcal{L}_{l \rightarrow v} \right].
\end{equation}
In practice, a symmetric cross-entropy loss is employed using the vision and language logits. The label $y\in\mathbb{R}$, which is the index of the image-text pair in the batch, represents the correspondence of the paired samples. For a batch of image--text pairs, $Y\in\mathbb{R}^b$ denotes the set of labels. The visual and textual logits, $\yv \in \mathbb{R}^{b \times b}$ and $\yt \in \mathbb{R}^{b \times b}$, are then calculated as:
\begin{equation}
\label{eq-clip-yhat}
    \yv =  \exp(\tau)\Ev\Et^\transpose \quad , \quad \yt =  \yv^\transpose,
\end{equation}
respectively. Ultimately, the overall CLIP loss is calculated using the cross-entropy loss, $H$, as:
\begin{equation}
\label{eq-clip-crossentropy}
    \mathcal{L}_{\textrm{clip}} = \frac{1}{2} [H( \yv, Y) + H( \yt, Y)].
\end{equation}

\noindent \textbf{Cross-Modal Alignment Score.} In contrastive representation learning, the goal is to learn similar representations for positive pairs and distant representations for irrelevant negative samples.  
In cross-modal vision--language learning, often the paired image--texts form the positive pair distribution.
Alignment entails mapping positive pairs to close embedding vectors such that a perfect alignment is achieved when $f(x_1) = f(x_2)$ for a given encoding function $f$ and a randomly drawn positive pair of samples $x_1$, $x_2$ \citep{understanding-align-uniform}. In this work, we are particularly interested in studying the alignment property for the CLIP embedding space as it represents the modality gap. We adopt the alignment measurement proposed by \citet{cyclip} and define it as the average cosine similarity between the positive pairs in the CLIP embedding space, i.e., the paired image and text embeddings:
\begin{equation}
\label{eq-alignment}
    \textrm{alignment} =  \frac{1}{N} \sum_{i=1}^{N} \evi \cdot \eti, \quad \quad \textrm{alignment} \in [-1, 1].
\end{equation}
Higher scores demonstrate better alignment, and, therefore, a decrease in the modality gap. 
\section{AlignCLIP}
\begin{figure}[t]
\centering
  \includegraphics[width=0.7\linewidth]{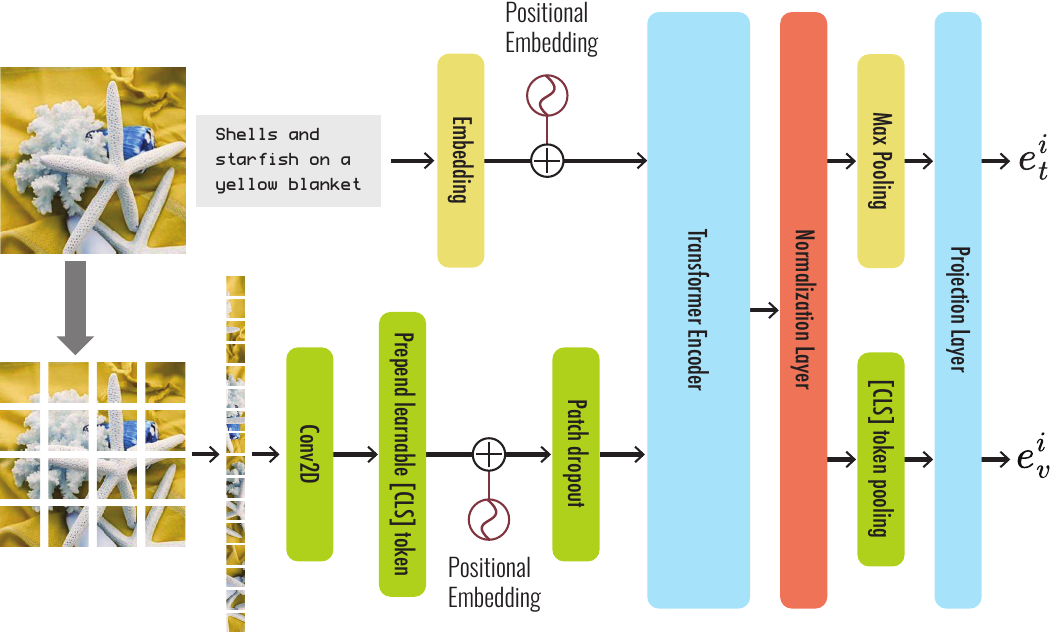}
  \caption{Overview of sharing the transformer and projection layer in SharedCLIP.}
  \label{fig-sharedCLIP}
\end{figure}
\subsection{Sharing the Learnable Parameter Space in CLIP}
For answering whether a shared parameter space reduces the gap, \myclip employs a transformer-based encoder architecture \citep{transformer}, where the transformer is shared between the two vision and language modality encoders. We suspect that one of the main reasons that the modality gap exists in the original CLIP's embedding space is the fact that each modality has a separate disentangled set of parameters for optimization. As a result, even with the same architecture for encoding each of the modalities, the final values of the learned parameters associated to each modality can diverge quite radically. This difference can lead to distinct functions that map each modality to completely different subregions of the embedding space.

Therefore, we seek to align the outputs of the vision and language encoding functions by sharing their parameter space to the extent possible. Specifically, we share the parameters of the transformer encoder as well as the projection layer between the vision and language modalities. We utilize a standard transformer encoder architecture \citep{vit}. An overview of our refined overall model architecture is given in Figure~\ref{fig-sharedCLIP}. Yellow components are designated for the language modality, green components for the visual modality, and the blue parts are shared between the two modalities. For encoding texts, the yellow and blue parts of the model actively contribute in the embedding calculations. Similarly, when encoding images, the green components as well as the blue ones are invoked for the computation. Following the original CLIP, we use max-pooling for text embeddings and $[CLS]$ token embedding for the image embedding. For simplicity, we refer to this architecture as \sharedclip throughout the rest of the paper. Sharing parameters in CLIP has been previously investigated from the perspective of downstream performance \citep{sharing_params}. In this work, we rather show its effectiveness with respect to the cross-modal alignment property.

\subsection{Intra-Modality Separation}
\label{method-imsep}
As shown in Figure~\ref{fig-intro-alignment}A, each modality resides in a distinct dense subregion of the CLIP embedding space. We hypothesize that this phenomenon is a direct result of the \textit{cross-modal} contrastive objective in CLIP, which merely optimizes the relative distances of image embeddings and text embeddings. The cross-modal contrastive loss alone is not sufficient for imposing meaningful distances within the uni-modal embeddings, i.e., pairwise distances of text embeddings and pairwise distances of image embeddings. Therefore, we define an objective function that enforces reasonable distances within each modality by separating the uni-modal embeddings that are semantically dissimilar. In other words, we impose a semantically-regularized Intra-Modality Separation (\imsep) in addition to the CLIP's objective function.

\imsep is achieved by a vision to vision contrastive loss, where image--text pairs are the positive samples and any pairwise combination of image--image is considered as negative samples, respectively:
\begin{equation}
\label{eq-contrastive-imsep}
   \mathcal{L}_{v \rightarrow v} = \frac{-1}{N} \sum_{i=1}^{N} \log \frac{\exp[(\evi \cdot \eti)\mathbin{/}\tau]}{\sum_{j=1, \atop{i \neq j}}^N \exp[(\evi \cdot \evj)\mathbin{/}\tau]}.
\end{equation}

In practice, given a batch of encoded image--text pairs, \imsep creates the vision-vision $\yvsep$ logits and then minimizes the cross-entropy loss over $Y$ and $\yvsep$.
To this end, first the pairwise cosine similarities of the images in the batch are calculated by:
\begin{equation}
\label{eq-ytt}
\Evv = \Ev\Ev^\transpose, \quad \text{where} \quad \Evv \in\mathbb{R}^{b \times b}.
\end{equation}

One should notice that while enforcing intra-modality separation, by minimizing the denominator in Eq.~\ref{eq-contrastive-imsep}, some samples might indeed be semantically similar to each other and therefore, must not be separated immensely. To this end, we regularize the intra-modality separation with respect to the pairwise semantic similarity of the samples. In order to calculate the semantic similarity within the image samples, we utilize their pairing texts as the semantic supervision signal, and invoke a pre-trained sentence encoder for encoding each text as $\esi \in \mathbb{R}^d$. 
We denote the corresponding batch of semantically encoded texts as $\Es \in \mathbb{R}^{b \times d}$, and proceed to calculate the pairwise semantic similarity $S \in \mathbb{R}^{b \times b}$ and the distance $D \in \mathbb{R}^{b \times b}$ of the texts:
\begin{equation}
\label{eq-S}
\mathcal{S} = \frac{\Es\Es^\transpose}{\norm {\Es}^2}, \quad \quad \mathcal{D} = 1 - \mathcal{S},
\end{equation}

respectively. We rely on $\mathcal{D}$ in order to re-scale $\Evv$. The goal of this re-scaling is to enforce a smaller dot product of the encoded images, if they are semantically similar, i.e., have a smaller distance in $D$. Conversely, the dot products in $\Evv$ obtain larger values when the text samples are semantically distant according to $D$. By this re-scaling, we seek to control the minimization of $(\evi \cdot \evj)$ when the samples are semantically similar. This is because the re-scaling mechanism enforces reducing these in-modality dot product values, and therefore, the dot products do not get strongly minimized in the cross-entropy loss, since the values are already small. Re-scaling is performed by:
\begin{equation}
\label{eq-rescale}
\Evs = \Evv \odot \mathcal{D}, \quad \text{where} \quad \Evs \in \mathbb{R}^{b \times b}
\end{equation}
\vspace*{-1mm}

where $\odot$ denotes the element-wise product. 

\begin{figure*}[t]
\centering
  \includegraphics[width=0.98\linewidth]{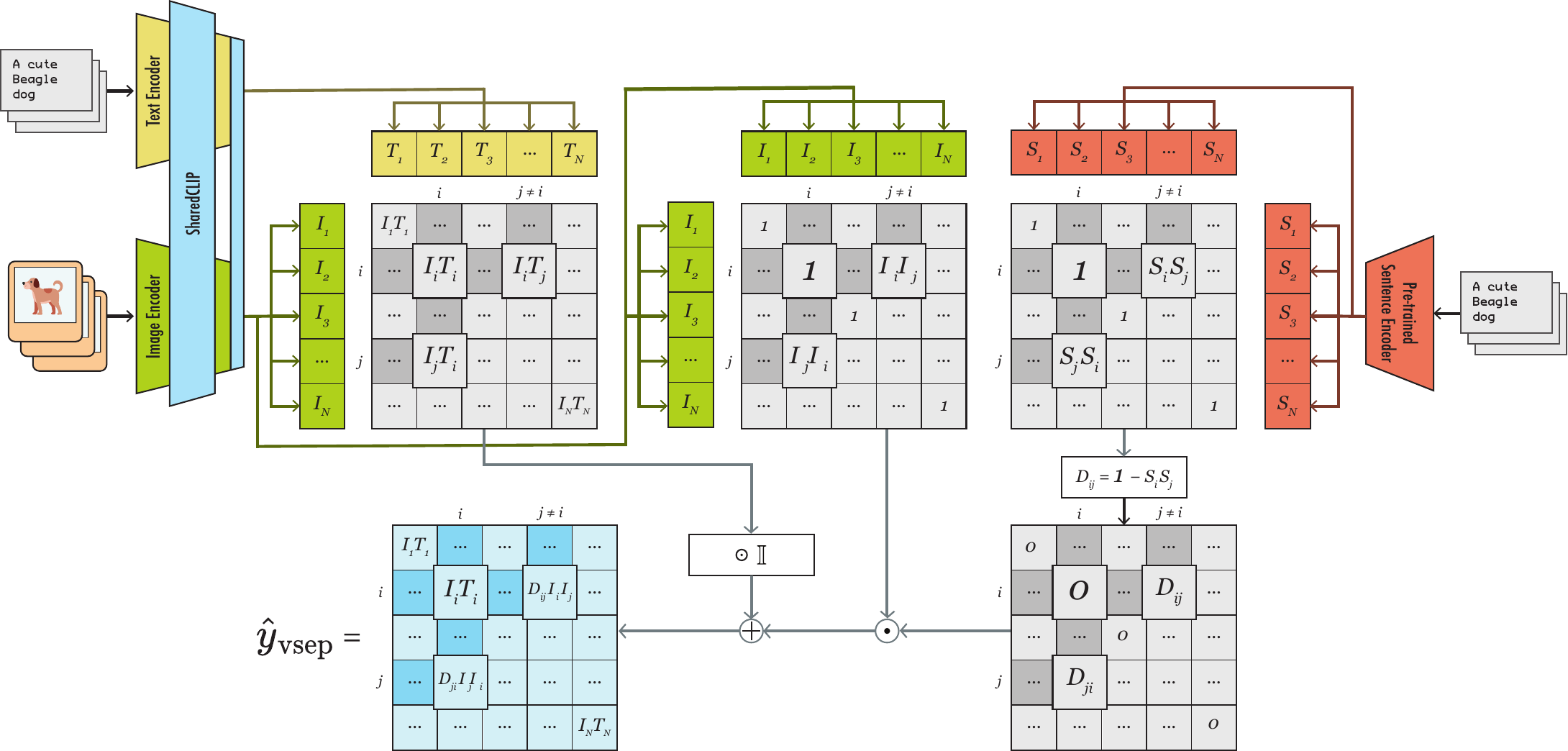}
  \caption{Schematic summary of the Intra-Modality Separation approach in \myclip.}
  \label{fig-imsep}
\end{figure*}

Afterwards, we calculate $\Evt=\Ev\Et^\transpose$ and mask the non-diagonal values by $\textrm{diag}(\Evt) = \mathbb{I} \odot \Evt$. We then obtain the vision--vision logits by:
\begin{equation}
\vspace*{-1mm}
\label{eq-yimsep}
\yvsep = \exp (\tau) \cdot [\text{diag}(\Evt) + \Evs], \quad \yvsep \in \mathbb{R}^{b \times b}
\end{equation}
In Figure~\ref{fig-imsep}, our approach for obtaining \imsep is schematically summarized.
Ultimately, we define the Intra-Modality Separation loss as:
\begin{equation}
\label{eq-imsep-loss}
\mathcal{L}_{\textrm{IMsep}} = H( \yvsep, Y),
\end{equation}
and adopt the core of the CLIP loss to represent the cross-modal separation:
\begin{equation}
\label{eq-crosssep-loss}
\mathcal{L}_{\textrm{CRsep}} = H( \yv, Y) + H( \yt, Y).
\end{equation}
The final loss function optimized in \myclip is:
\begin{equation}
\label{eq-total-loss}
\mathcal{L} = \alpha \mathcal{L}_{\textrm{CRsep}} + \beta \mathcal{L}_{\textrm{IMsep}}
\end{equation}

Note that it is sufficient to define the Intra-Modality Separation function for only one of the modalities, image in our case, since the cross-modal objective defined by $\mathcal{L}_{\textrm{CRsep}}$ already behaves as a supervision for the other modality, text in our case, and enforces the Intra-Modality Separation in that modality as well. To better support this statement, we extensively experiment the effects of adding the similar Intra-Modality Separation on text embeddings in Section~\ref{appendix-text-separation} and show that it is sufficient to impose the Intra-Modality Separation on the image embeddings.
\section{Experiments}
\label{sec-exp}
\subsection{Training Dataset and Setup}
\label{exp-training-setup}
We used the Conceptual Caption 12M (\trainingdata) dataset, which has been similarly used in previous work \citep{cyclip, slip, supervision-exists, cc12m}, for pre-training the models. 
We set our setup quite similar to that of the original CLIP with the ViT-B-16 backend, in order to ensure a fair comparison of CLIP with \sharedclip and \myclip. The pre-trained semantic encoder utilized in \myclip for re-scaling image--image cosine similarities is the SBERT all-mpnet-base-v2 model. 
In order to fairly compare the effectiveness of each model, we trained all of them  
from scratch using the \trainingdata dataset and the OpenCLIP implementation \citep{openclip, openclip-software}. Each model was trained for 6 days using an NVIDIA H100 GPU with batch size 512 for 30 epochs using AdamW optimization with a starting learning rate of $1 \times 10^{-3}$, cosine scheduler, 10,000 warmup steps, a weight decay of $0.1$, and an initial temperature value of 0.07. The output embedding dimension for both the vision and language modalities is set to 768. We used the checkpoint from the last epoch in our evaluations of downstream experiments. In \myclip, we set $\alpha=1$ and $\beta=\frac{1}{2}$. In Section~\ref{appendix-setup}, more details about the training setup is provided.

Since our goal is to shed lights on a specific shortcoming of the original CLIP model, i.e., the modality gap problem, we compare our results to the original CLIP model in this paper. Further comparisons to other state-of-the-art models is not the focus of this study, as the goal is to investigate modifications that reduce the modality gap in CLIP without losing performance in downstream tasks, rather than achieving the state-of-the-art results.

\begin{table}[t]
\vskip 0.15in
\begin{center}
\begin{small}
\begin{sc}
\resizebox{0.9\linewidth}{!}{
\begin{tabular}{lccccc}
\toprule
\textbf{Model} & \textbf{CC3M} & \textbf{MSCOCO} & \textbf{ImageNet-1K} & \textbf{CIFAR-100} & \textbf{CIFAR-10}\\
\midrule
CLIP    & 0.42 & 0.47 & 0.41 &0.38&0.4\\
\midrule
SharedCLIP    & 0.59  & 0.62  & 0.57 &0.54&0.54\\
\midrule
AlignCLIP & \textbf{0.64} & \textbf{0.67} & \textbf{0.63} & \textbf{0.62} & \textbf{0.64}\\
\bottomrule
\end{tabular}
}
\end{sc}
\end{small}
\end{center}
\caption{Comparison of the alignment score.}
\label{res-table-alignment}
\end{table}
\begin{figure}[t]
\centering
  \includegraphics[width=\linewidth]{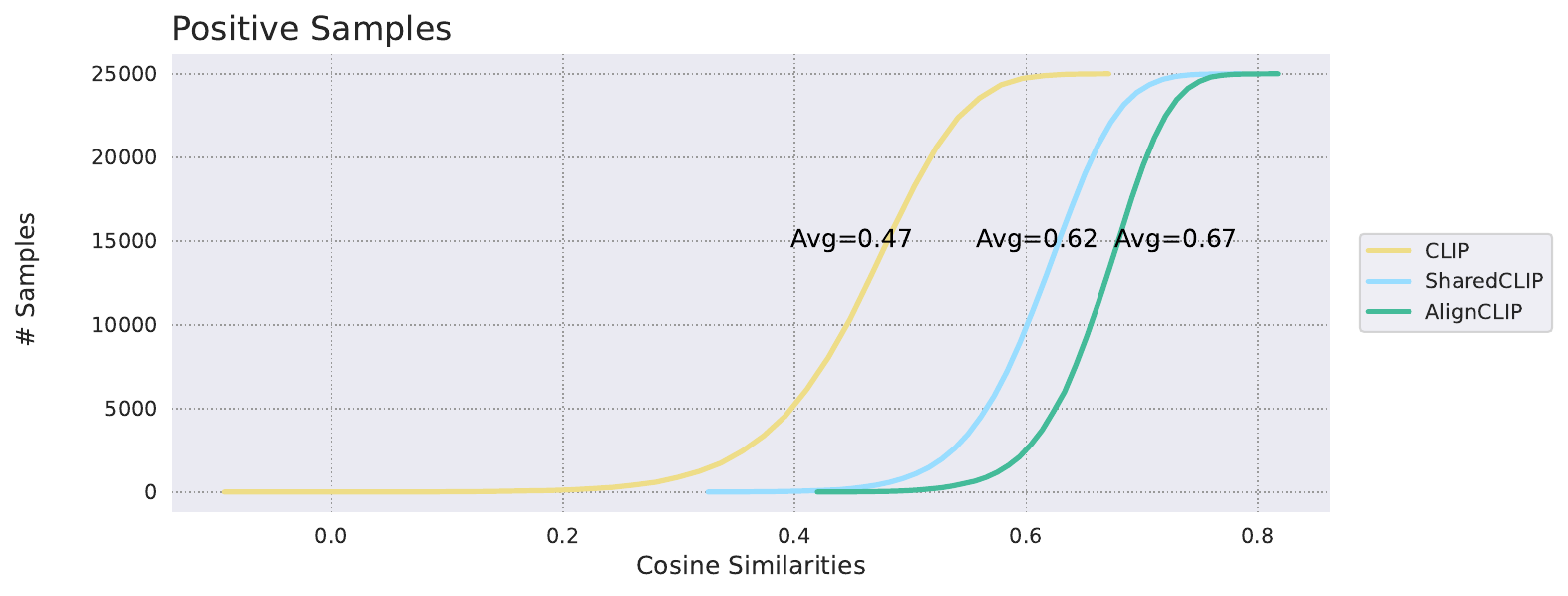}
  \caption{Cumulative distribution of pairwise cosine similarities of positive samples in MSCOCO.}
  \label{fig-cosine-positives}
\end{figure}
\subsection{Cross-Modal Alignment} 
We start by reporting and comparing the alignment scores when using CLIP, SharedCLIP, and AlignCLIP models on the validation sets from CC3M, MSCOCO as well as the ImageNet-1K, CIFAR-100, and CIFAR-10 test datasets. Table~\ref{res-table-alignment} summarizes the corresponding alignment scores. We observe that the original CLIP model has relatively low alignment scores, varying within $[0.38, 0.47]$, across all five datasets. These scores mean that the average angle between the paired image--text embeddings is a value between 61 and 68 degrees. In contrast, sharing the parameter space in \sharedclip results in noticeable improvements of up to 0.17 in the alignment scores. As a result, the average angle between the paired image--text embeddings decreases to about 51 degrees. Furthermore, using \myclip yields even better alignment scores, ranging from 0.62 to 0.67, and a decreased average angle of 47 degrees between the multi-modal paired samples. These observations confirm that \myclip improves the cross-modal alignment in CLIP and thereby, reduces the modality gap. 

We also plot the cumulative distribution of cosine similarities of the positive samples from MSCOCO validation dataset when encoded using the original CLIP, \sharedclip, and \myclip in Figure~\ref{fig-cosine-positives}. We find that using \sharedclip noticeably shifts the distribution of the cosine similarity of positive samples towards higher similarity values. A higher similarity of positive samples, i.e., image--text pairs, means achieving greater cross-modal alignment, and thus a lower modality gap. Furthermore, \myclip shifts the distribution even more to the right, resulting in higher similarities of positive samples, better cross-modal alignment, and a reduction of the modality gap.

\begin{table}[t]
\vskip 0.15in
\begin{center}
\begin{small}
\begin{sc}
\resizebox{\linewidth}{!}{
\begin{tabular}{lcccccccccc}
\toprule
\textbf{Model} & \multicolumn{2}{c}{\textbf{ImageNet-1K}} & \multicolumn{2}{c}{\textbf{CIFAR-100}} & \multicolumn{2}{c}{\textbf{CIFAR-10}} & \multicolumn{2}{c}{\textbf{Flowers-102}}& \multicolumn{2}{c}{\textbf{Stanford Cars}}\\
\cmidrule(l){2-3}
\cmidrule(l){4-5}
\cmidrule(l){6-7}
\cmidrule(l){8-9}
\cmidrule(l){10-11}
& Top1 & Top5& Top1& Top5& Top1 & Top5 & Top1 & Top5& Top1 & Top5\\
\midrule
CLIP    & 31.4 & 58.7 & 28.1 & 55.9 & 61.5 & 95.6 &18&39.1 & 11.6 & 36.5\\
\midrule
SharedCLIP   &  32.1 & 59.7 & 26.4 & 54.7 & 56.9 & 95.2 &18.2&38.9&10.7&35.4\\
\midrule
AlignCLIP & \textbf{32.8} & \textbf{60.6} &  \textbf{36.5} & \textbf{66.4} & \textbf{69.3} & \textbf{97.8} & \textbf{18.8}& \textbf{40.3} &\textbf{11.8}&\textbf{38.1}\\
\bottomrule
\end{tabular}
}
\end{sc}
\end{small}
\end{center}
\caption{Accuracy scores for zero-shot image classification.}
\label{res-table-zeroshot-cls}
\end{table}
\begin{table}[h!]
\vskip 0.15in
\begin{center}
\begin{small}
\begin{sc}
\resizebox{\linewidth}{!}{
\begin{tabular}{lcccccc}
\toprule
\textbf{Model} & \textbf{ImageNet-1K} &\textbf{CIFAR-100} & \textbf{CIFAR-10} & \textbf{Flowers-102} & \textbf{Stanford Cars}\\
\midrule
CLIP   & 50 & 62.6 & 85 & 71.5 & 42.2\\
\midrule
SharedCLIP   &  51.2  & 63 & 85 & 74.4& 40.5\\
\midrule
AlignCLIP & \textbf{51.5} & \textbf{67.4} & \textbf{87.2} & \textbf{76.8} & \textbf{45.6}\\
\bottomrule
\end{tabular}
}
\end{sc}
\end{small}
\end{center}
\caption{Accuracy scores for image classification with linear probing.}
\label{res-table-linear-probing}
\end{table}
\subsection{Classification}
CLIP's pre-training objective for predicting whether a text is paired with an image has resulted in outstanding image classification performance when tested in a zero-shot setting as well as after linear probing \citep{clip}. Therefore, we further assess how sharing the learnable parameters and the intra-modality loss affect the classification performance in these settings.

\vspace{1mm}
\noindent\textbf{Zero-Shot Image Classification.} We conduct the zero-shot classification experiments on ImageNet-1K \citep{imagenet1k}, CIFAR-100, CIFAR-10 \citep{cifar}, Flowers-102 \citep{flowers102}, and Stanford Cars \citep{stanfordcars} with the combination of text prompts used by CLIP \citep{clip}, e.g., ``a photo of the \{label\}''. The experimental results summarized in Table~\ref{res-table-zeroshot-cls} show that \sharedclip reduces the accuracy of the zero-shot classification on CIFAR-10 by about $5\%$ and $0.4\%$ when measuring Top-1 and Top-5 accuracy scores, respectively. Similarly, \sharedclip's results on CIFAR-100 show about $2\%$ and $1\%$ decrease in accuracy. The trend of accuracy reduction is also observed on Flowers-102 and Stanford Cars datasets. However, on ImageNet-1K, \sharedclip evinces up to $1\%$ improvement of accuracy. 
In contrast, \myclip yields the best scores across all five datasets. It achieves $1.4\%$ and $2\%$ improvement of Top1 and Top5 accuracy, respectively, on ImageNet-1K when compared to the original CLIP. On CIFAR-10, \myclip achieves up to $8\%$ and $2\%$ improvements in Top-1 and Top-5 accuracy scores. Similarly, using \myclip for CIFAR-100 results in about $8\%$ and $11\%$ enhancement in Top-1 and Top-5 accuracy scores in comparison to the original CLIP. The trend of improvement is also observed on the Flowers-102 and Stanford Cars datasets. Our experiments thus evince that via sharing parameters and the additional intra-modality separation, \myclip improves the cross-modal alignment of the embeddings while enhancing the performance on the downstream zero-shot image classification task.

\noindent\textbf{Linear Probing.}
We further test the performance of \sharedclip and \myclip when performing linear probing for image classification and report the Top1 accuracy results in Table~\ref{res-table-linear-probing}. For all datasets, we train the linear classifier layer with a batch size of 128, for 30 epochs, with AdamW optimizer, and a cosine scheduler with a starting learning rate of 5e-4. Table~\ref{res-table-linear-probing} shows the superiority of \myclip in the task of image classification with linear probing across all 5 datasets.

\subsection{Robustness to Natural Distribution Shift}
In the zero-shot image classification task, CLIP has additionally shown impressive robustness to natural distribution shifts and promising generalizability to out-of-distribution images. Therefore, we expand our evaluations and investigate to what extent \sharedclip and \myclip change the performance with natural distribution shifts. We use the ImageNetV2, ImageNet-R, ImageNet-A, and ImageNetSketch datasets for these evaluations and report the corresponding results in Table~\ref{res-table-distr-shift}, in terms of Top-1 and Top-5 accuracy. It is first observed that \sharedclip generally improves the classification accuracy in comparison to the CLIP model. Secondly, \myclip achieves the best classification results across all datasets. In summary, when comparing to the CLIP model, \myclip achieves about $2\%$ and $1\%$ improvement in Top1 and Top5 accuracy on the ImageNetV2 dataset. On ImageNet-R, \myclip improves both Top1 and Top5 scores by about $2\%$. The positive general trend of the accuracy enhancement is also observed on the ImageNet-A and ImageNetSketch datasets. Based on these observations, we conclude that sharing parameters and applying intra-modality separation in \sharedclip and \myclip improves the robustness to natural distribution shifts when compared to the original CLIP model and at the same time, improve the modality gap.

\begin{table}[t]
\vskip 0.15in
\begin{center}
\begin{small}
\begin{sc}
\resizebox{0.8\linewidth}{!}{
\begin{tabular}{lcccccccc}
\toprule
\textbf{Model} & \multicolumn{2}{c}{\textbf{ImageNetV2}} & \multicolumn{2}{c}{\textbf{ImageNet-R}} & \multicolumn{2}{c}{\textbf{ImageNet-A}} & \multicolumn{2}{c}{\textbf{ImageNetSketch}} \\
\cmidrule(l){2-3}
\cmidrule(l){4-5}
\cmidrule(l){6-7}
\cmidrule(l){8-9}
& Top1 & Top5& Top1& Top5&Top1& Top5& Top1 & Top5 \\
\midrule
CLIP    & 27.1 & 53.3& 39.8 & 65.8& 6.5& 25.4 &19.4& 41.8\\
\midrule
SharedCLIP    & 27.5  & 53.5&  40.2  &  \textbf{67.3} & 6.7 &  25.5 & 20.6 &  \textbf{43.2} \\
\midrule
AlignCLIP & \textbf{29.1} & \textbf{54.4} & \textbf{41.2} & \textbf{67.3} & \textbf{7} & \textbf{25.6}  & \textbf{20.7} & \textbf{43.2}\\
\bottomrule
\end{tabular}
}
\end{sc}
\end{small}
\end{center}
\caption{Accuracy scores for zero-shot classification and natural distribution shift.}
\label{res-table-distr-shift}
\end{table}

\subsection{Multi-Modal Retrieval}
\vspace{1mm}
\noindent\textbf{Zero-Shot Transfer.} In addition to classification, we evaluate \sharedclip and \myclip in the application of zero-shot image-to-text and text-to-image retrieval using the validation splits from the MSCOCO \citep{mscoco} and Flickr30K \citep{flickr} datasets. The results of these evaluations are reported in Table~\ref{res-table-zeroshot-IR}. In all settings, the text prompt ``a photo of the \{caption\}'' is used. Our experiments show that both \sharedclip and \myclip improve the retrieval results measured by $R@\{1, 5, 10\}$ on both datasets when compared to the original CLIP model. In addition, \myclip achieves the best overall results in comparison to \sharedclip. When testing text-to-image retrieval on the MSCOCO dataset, \sharedclip outperforms \myclip at R@10. Similarly, for the image-to-text retrieval on the Flickr dataset, \sharedclip achieves the best results. These experiments demonstrate that it is possible to reduce the modality gap in CLIP via parameter sharing while improving the downstream multi-modal retrieval tasks. Furthermore, the addition intra-modality separation improves the alignment noticeably while noticeably enhancing the retrieval performance.

\begin{table*}[t]
\vskip 0.15in
\begin{center}
\begin{small}
\begin{sc}
\resizebox{\linewidth}{!}{
\begin{tabular}{lcccccccccccc}
\toprule
 & \multicolumn{6}{c}{\textbf{MSCOCO}} & \multicolumn{6}{c}{\textbf{Flickr30K}} \\
\cmidrule(l){2-7}
\cmidrule(l){8-13}
\textbf{Model} & \multicolumn{3}{c}{\textbf{I $\rightarrow$ T}} & \multicolumn{3}{c}{\textbf{T $\rightarrow$ I}} & \multicolumn{3}{c}{\textbf{I $\rightarrow$ T}} & \multicolumn{3}{c}{\textbf{T $\rightarrow$ I}} \\
\cmidrule(l){2-4}
\cmidrule(l){5-7}
\cmidrule(l){8-10}
\cmidrule(l){11-13}
& R@1 & R@5 & R@10 & R@1 & R@5 & R@10 & R@1 & R@5 & R@10 & R@1 & R@5 & R@10 \\
\specialrule{1.5pt}{1pt}{1pt}
CLIP & 31.4 & 57 & 68.6 & 20.5 & 44.1 & 55.9 & 53.2 & 80.5& 88.6& 39.9& 69& 78.5\\
\midrule
SharedCLIP &  33.5  &  59.6  &  70.8  &  21.8  &  \textbf{45.4}  &  \textbf{57.3}  &  \textbf{58.3}  &  \textbf{83.6}  & \textbf{89.8} &  42.5 &  70 &  \textbf{79.1} \\
\midrule
AlignCLIP & \textbf{35.1} & \textbf{60.8} & \textbf{71.4} & \textbf{21.9} & \textbf{45.4} & 56.8 & 57.2 & 82.3 & \textbf{89.8} & \textbf{42.7} & \textbf{70.2}  & \textbf{79.1}\\
\bottomrule
\end{tabular}
}
\end{sc}
\end{small}
\end{center}
\caption{Zero-shot cross-modal retrieval summarized with R@\{1, 5, 10\}.}
\label{res-table-zeroshot-IR}
\end{table*}
\begin{table*}[t]
\vskip 0.15in
\begin{center}
\begin{small}
\begin{sc}
\resizebox{\linewidth}{!}{
\begin{tabular}{lcccccccccccc}
\toprule
 & \multicolumn{6}{c}{\textbf{MSCOCO}} & \multicolumn{6}{c}{\textbf{Flickr30K}} \\
\cmidrule(l){2-7}
\cmidrule(l){8-13}
\textbf{Model} & \multicolumn{3}{c}{\textbf{I $\rightarrow$ T}} & \multicolumn{3}{c}{\textbf{T $\rightarrow$ I}} & \multicolumn{3}{c}{\textbf{I $\rightarrow$ T}} & \multicolumn{3}{c}{\textbf{T $\rightarrow$ I}} \\
\cmidrule(l){2-4}
\cmidrule(l){5-7}
\cmidrule(l){8-10}
\cmidrule(l){11-13}
& R@1 & R@5 & R@10 & R@1 & R@5 & R@10 & R@1 & R@5 & R@10 & R@1 & R@5 & R@10 \\
\specialrule{1.5pt}{1pt}{1pt}
CLIP & 39.6 & 67.5  & 78.3&  26.7 & 53.4 & 65.7& 64.8& 87.8& 93.9& 47.4& 75.9 & 84\\
\midrule
SharedCLIP & 40.7 & 69.2  & 79.6& \textbf{27.9} & \textbf{55} & \textbf{66.7} & \textbf{66.5} & \textbf{89.1}& 94.1& 48.9 & 76.4 & 84.3\\
\midrule
AlignCLIP & \textbf{41.7} & \textbf{69.3}  & \textbf{80.1}& \textbf{27.9} & 54.8 & 66.2 & 66 & \textbf{89.1}& \textbf{94.5}& \textbf{49.4}& \textbf{76.7} & \textbf{84.4}\\
\bottomrule
\end{tabular}
}
\end{sc}
\end{small}
\end{center}
\caption{Fine-tuned cross-modal retrieval summarized with R@\{1, 5, 10\}.}
\label{res-table-finetuned-IR}
\end{table*}
\vspace{1mm}
\noindent\textbf{Fine-tuning Multi-Modal Retrieval.} Table~\ref{res-table-finetuned-IR} summarizes the result of multi-modal retrieval when each model is fine-tuned using the corresponding training set. We fine-tuned each model for 8 and 20 epochs on MSCOCO and Flickr, respectively. In both cases, the batch size was set to 128 and the AdamW optimizer with the learning rate of 5e-6 and a weight decay of 0.2 was used. Our results show that both \sharedclip and \myclip outperform the CLIP model in the fine-tuning scenario. Additionally, \myclip generally achieves a slightly better performance in comparison to \sharedclip.

\subsection{Ablation Study}
This section provides an ablation study on the effectiveness of the re-scaling mechanism proposed in Eq.~\ref{eq-rescale}. In order to see the impacts on the different types of downstream tasks, i.e., image classification, classification with distribution shift and multi-modal retrieval, we compare the performance of \myclip with and without the re-scaling mechanism on ImageNet-1K, CIFAR-100, CIFAR-10, ImageNetV2, MSCOCO and Flickr30K datasets and summarize the results in Table~\ref{res-table-ablation}. This study substantially concludes that the re-scaling mechanism is effective in controlling the separation of similar image samples in the batch and therefore, increasing the results in the downstream tasks.
\begin{table}[h!]
\vskip 0.15in
\begin{center}
\begin{small}
\begin{sc}
\resizebox{\linewidth}{!}{
\begin{tabular}{lcccccccc}
\toprule
\textbf{Model} & \textbf{ImageNet-1K} &\textbf{CIFAR-100} & \textbf{CIFAR-10} &\textbf{ImageNetV2}  &\multicolumn{2}{c}{\textbf{MSCOCO}} & \multicolumn{2}{c}{\textbf{FLICKR}}\\
\cmidrule(l){2-2}
\cmidrule(l){3-3}
\cmidrule(l){4-4}
\cmidrule(l){5-6}
\cmidrule(l){7-8}
\cmidrule(l){9-9}
&Top1&Top1& Top1 & Top1 & I $\rightarrow$ T & T $\rightarrow$ I & I $\rightarrow$ T & T $\rightarrow$ I \\

\midrule
AlignCLIP-W/O Rescaling   &  32.8  & 34.7 & 64.2  & 27.9 & 34 & 59.7 & 56 & 42.5 \\
\midrule
AlignCLIP   & 32.8 & \textbf{36.5} & \textbf{69.3}  & \textbf{29.1} & \textbf{35.1} & \textbf{60.8} & \textbf{57.2} & \textbf{42.7}\\
\bottomrule
\end{tabular}
}
\end{sc}
\end{small}
\end{center}
\caption{Ablation study on the re-scaling mechanism in AlignCLIP.}
\label{res-table-ablation}
\end{table}
\subsection{Results Analysis} 
\begin{figure}[t]
     \centering
     \begin{subfigure}[b]{0.32\textwidth}
         \centering
         \includegraphics[width=\textwidth]{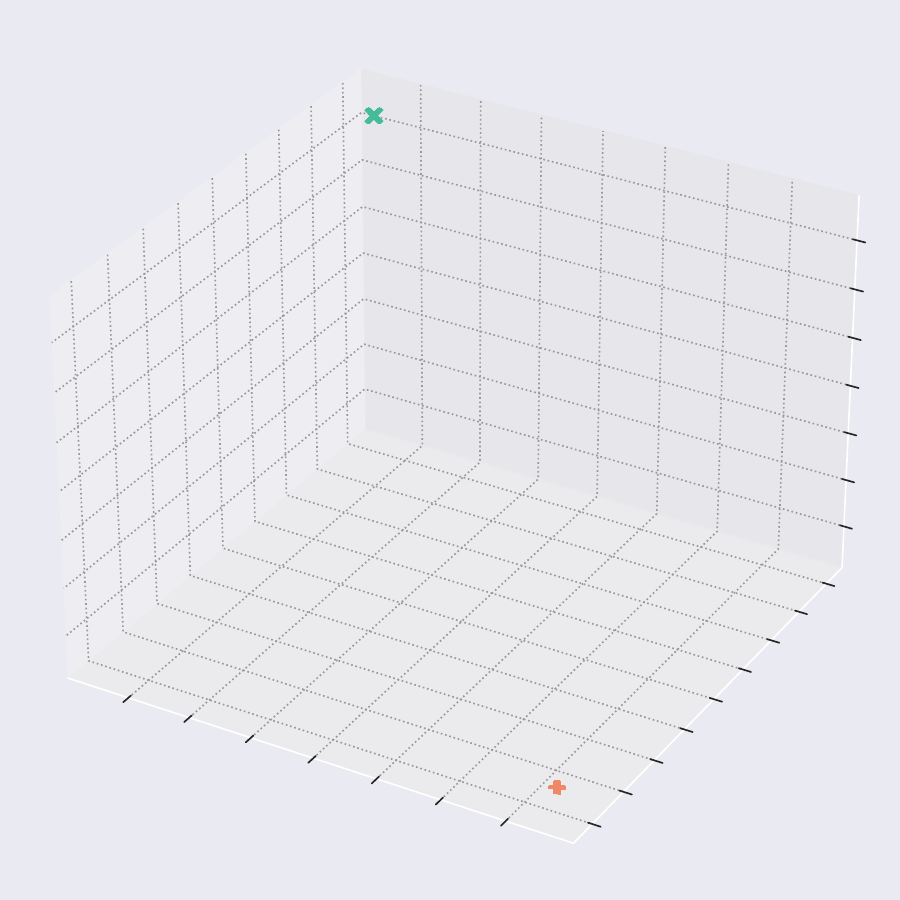}
         \caption{Original CLIP}
         \label{fig:dosnes-clip}
     \end{subfigure}
     \hfill
     \begin{subfigure}[b]{0.32\textwidth}
         \centering
         \includegraphics[width=\textwidth]{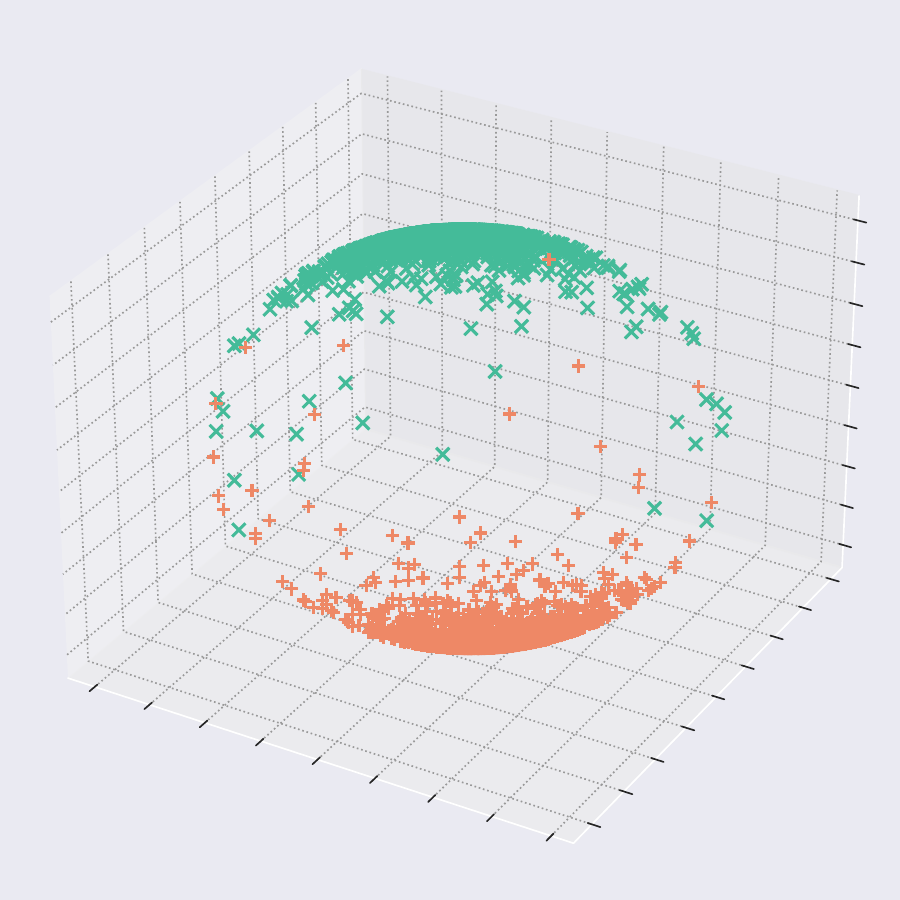}
         \caption{SharedCLIP}
         \label{fig:dosnes-sharedclip}
     \end{subfigure}
     \hfill
     \begin{subfigure}[b]{0.32\textwidth}
         \centering
         \includegraphics[width=\textwidth]{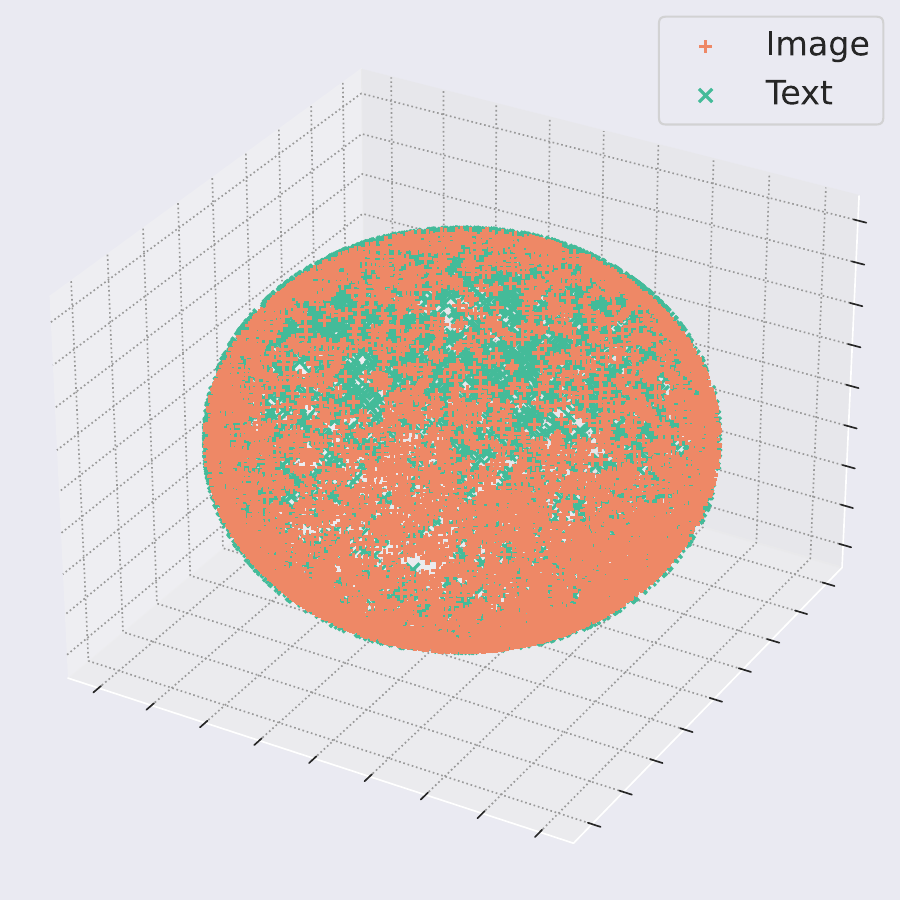}
         \caption{AlignCLIP}
         \label{fig:dosnes-alignclip}
     \end{subfigure}
        \caption{DOSNES visualization of the multi-modal embeddings using CC3M}
        \label{fig:dosnes}
\end{figure}

\vspace{1mm}
\noindent\textbf{DOSNES Visualization.} For a more comprehensive comparison of the distribution of each modality, we visualize the DOSNES projection of the encoded image--texts from the MSCOCO validation set in Figure~\ref{fig:dosnes}. 
As can be seen, the uni-modal embeddings in CLIP are densely located on opposite sides of the hypersphere. In contrast, the embeddings get spread out when using the \sharedclip model. Finally, the embedding space of \myclip achieves the best spread of the uni-modal embeddings and substantially greater alignment of image and text embeddings. Thus, the intra-modality separation leads to a better alignment and a substantial reduction of the modality gap. Furthermore, Figure~\ref{fig:dosnes-alignclip} shows that \myclip reduces the sparsity of the embeddings on the hypersphere.

\vspace{1mm}
\noindent\textbf{Comparison of Qualitative Examples.} In Table~\ref{res-table-mscoco-example}, examples from the MSCOCO validation dataset is provided where the images convey the same general semantics but one of the ground truth texts provides more detailed information. On the left side of Table~\ref{res-table-mscoco-example}, two images with their corresponding ground truth captions are provided. We use CLIP, \sharedclip and \myclip for encoding the images and texts, and provide the cosine similarities on the right side of the table. As can be seen, when querying the first image, the similarity of the first image and the second text using the CLIP embeddings is higher in comparison to the ground truth caption. Suggesting that the second text will get selected as the predicted caption when using CLIP. This flaw still appears when using \sharedclip for encoding the images and texts. However, when using \myclip, the cosine similarity of the first image and the first text is higher in comparison to the second text, meaning that when querying the first image, the ground truth caption, which is semantically more correct in comparison to the second text, successfully gets selected. This suggests that the semantic regularization of the Intra-Modality Separation in \myclip, which is calculated using the semantics of the text samples, potentially contributes in improving the retrieval performance.

\begin{table}[t]
\vskip 0.15in
\begin{center}
\begin{small}
\begin{sc}
\resizebox{\linewidth}{!}{
\begin{tabular}{lccccccc}
\toprule
\multirow{4}{*}{\raisebox{-0.8\totalheight}{\includegraphics[width=0.75\textwidth]{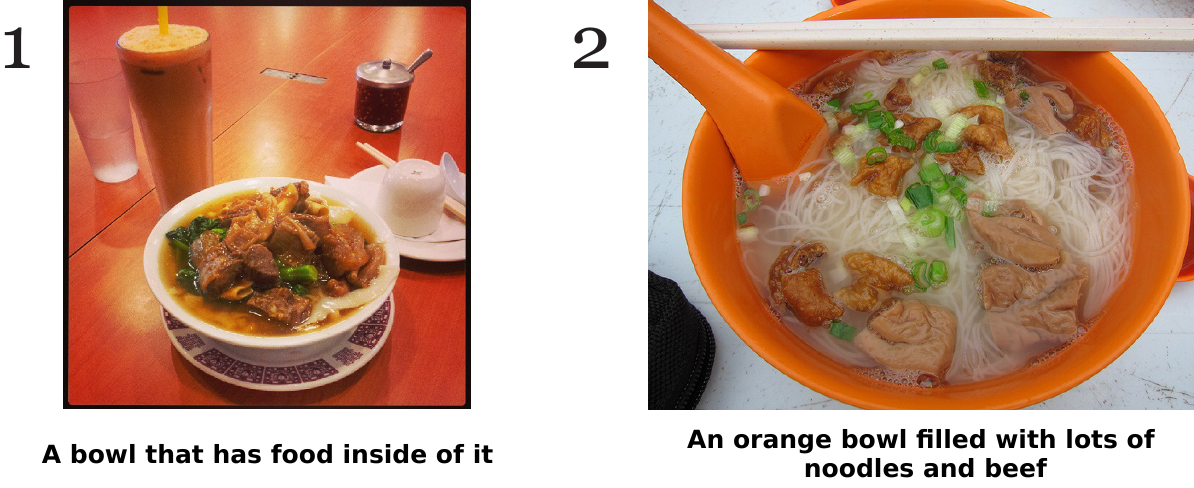}}} 
&&&&&&&\\
 & & \multicolumn{2}{c}{\rotatebox{45}{\textbf{CLIP}}} & \multicolumn{2}{c}{\rotatebox{45}{\textbf{SharedCLIP}}} & \multicolumn{2}{c}{\rotatebox{45}{\textbf{AlignCLIP}}} \\
\cmidrule(l){3-4}
\cmidrule(l){5-6}
\cmidrule(l){7-8}
     & &$T_1$ & $T_2$ & $T_1$ & $T_2$ & $T_1$ & $T_2$ \\
\cmidrule(l){3-8}
   & $I_1$ &0.37   & 0.42   & 0.53   & 0.55   & 0.61   & 0.58 \\
   & $I_2$ & 0.39  & 0.52   & 0.61   & 0.64   & 0.63   & 0.71 \\
   &&&&&&&\\
   &&&&&&&\\
   &&&&&&&\\  
\bottomrule
\end{tabular}
}
\end{sc}
\end{small}
\end{center}
\vskip 0.1in
\caption{Qualitative example of semantically similar samples from MSCOCO.}
\label{res-table-mscoco-example}
\end{table}
\section{Related Work}
\label{section-related}
\textbf{Modifications of CLIP.} Prior work has sought to improve several different aspects of CLIP. UniCLIP \citep{uniclip} proposed a multi-positive contrastive loss for training augmentation-aware feature embeddings. SLIP \citep{slip} added an additional image-based contrastive objective function using augmented images as positive samples to the original CLIP's loss function. Furthermore, CyCLIP~\citep{cyclip} formalized the geometric consistency of the image and text embeddings in the CLIP embedding space and added a loss function for regularizing the cross-modal and in-modal similarity scores.  
The authors provide an analysis of the cross-modal alignment showing that the final CyCLIP model does not improve the alignment of image--text embeddings. 
In xCLIP \citep{xclip}, a non-contrastive training regimen based on image--text pairs is adopted in order to mitigate the requirement of large batch sizes in contrastive learning.  
With a similar goal, SigLIP \citep{siglip} employed a pairwise sigmoid loss for scaling up the batch size.  
The recently published ReCLIP \citep{reclip} had the goal of refining the fine-tuning of CLIP in domain adaptation settings, based on unlabeled target samples.  
While ReCLIP results in noticeable gains for the considered classification task, its effectiveness with regard to the cross-modal alignment property is not studied by the authors. 
In contrast to previous work, \myclip studies the cross-modal alignment and modality gap phenomena in CLIP by investigating the effects of parameter-sharing as well as intra-modality separation.

\textbf{Modality Gap in CLIP.} The modality gap in CLIP's embedding space was first studied by \citet{mind_gap}. Authors showed that the modality gap is caused by a combination of the model initialization and the contrastive loss optimization. Furthermore, they showed that CLIP's embedding space is very sparse such that the effective embedding space is an extremely narrow cone. In a similar study, \citet{isotropy} measured the alignment of image and text embeddings in CLIP with a focus on the isotropic properties. More recently, \citet{two_effects} showed that one of the key factors contributing in both the modality gap and the bias in CLIP is the information-imbalance between the two modalities, i.e., images often have much more detailed information of the scene in comparison to their corresponding text descriptions. Furthermore, \citet{understanding_structure} showed that even under the perfect alignment, the prediction error in downstream tasks cannot be smaller than the information gap that exists between the modalities. They further propose intra-modality as well as inter-modality regularization using augmented samples in order to improve the latent embeddings structures. In contrast to the previous work, we study the effectiveness of parameter sharing between the modality encoders as well as semantically-regularized separation of the uni-modal embeddings on reducing the modality gap, under the same amount of information gap across modalities.
\section{Conclusion}
\label{sec-conclusion}
This work investigates the potential of reducing the modality gap in CLIP by sharing the learnable parameter space of the vision and language encoders as well as enforcing a contrastive intra-modality separation objective. We further examine the effects of modality gap reduction by the aforementioned refinements in the performance of downstream tasks.  
Through extensive experiments, we show that \sharedclip and \myclip improve various zero-shot as well as fine-tuning downstream applications when compared to CLIP, while substantially improving the cross-modal alignment, and therefore, reducing the gap. Our work shows that it is possible to mitigate the modality gap in CLIP via parameter sharing and intra-modality separation without losing downstream performance.

\section*{Acknowledgments}
We acknowledge the financial support from the German Federal Ministry for Education and Research (BMBF) within the project KI-Servicezentrum Berlin Brandenburg (01IS22092).

\bibliographystyle{plainnat}
\bibliography{neurips_2024}
\clearpage
\appendix
\section{Appendix / Supplemental Material}
\label{appendix}
\subsection{Training Dataset and Setup}
\label{appendix-setup}
For pre-training, we used the Conceptual Caption 12M (\trainingdata) dataset \citep{cc12m}. In comparison to the data used for pre-training CLIP, i.e., 400M image--text pairs, \trainingdata is a much smaller dataset, with about 12M noisy image--text pairs. Therefore, our results are not directly comparable to the numbers reported in the original CLIP paper \citep{clip}. Nonetheless, \trainingdata is one of the popularly used large-scale and publicly available datasets that enables pre-training vision--language models and analyzing the effectiveness of different training paradigms on benchmark evaluations.  

We adopted a transformer encoder consisting of 12 layers and 12 heads in \sharedclip and \myclip. The same input pre-processing and augmentations employed in CLIP have been used for \sharedclip as well as \myclip, including random cropping of images to the size $224 \times 224$. The image patch size for encoding visual data is set to $16\times16$. When encoding texts, the maximum sequence length is set to 77 tokens and the vocabulary size for the embedding layer is set to 49,408. The output embedding dimension for both the vision and language modalities is set to 768. We chose our setup to be quite similar to that of the original CLIP with the ViT-B-16 backend, in order to ensure a fair comparison of CLIP and our proposed model. SBERT all-mpnet-base-v2 model has been utilized in \myclip for re-scaling text--text and image--image cosine similarities is the .

In order to fairly compare the effectiveness of each model, we trained all of the models , i.e., the original CLIP with ViT-B-16 backend, SharedCLIP, and AlignCLIP, 
from scratch using the \trainingdata dataset and the OpenCLIP implementation \citep{openclip, openclip-software}. For all models, we used AdamW optimization with a starting learning rate of $1 \times 10^{-3}$, cosine scheduler, 10,000 warmup steps, and a weight decay of $0.1$. The initial temperature value for all models were set to 0.07. Each model was trained using an NVIDIA H100 GPU with batch size 512 for 30 epochs. We used the checkpoint from the last epoch in our evaluations of downstream experiments. In \myclip, we set $\alpha=1$ and $\beta=\frac{1}{2}$.

\begin{table}[t]
\vskip 0.15in
\begin{center}
\begin{small}
\begin{sc}
\resizebox{\linewidth}{!}{
\begin{tabular}{lcccccccccc}
\toprule
\textbf{Model} & \multicolumn{2}{c}{\textbf{ImageNet-1K}} & \multicolumn{2}{c}{\textbf{CIFAR-100}} & \multicolumn{2}{c}{\textbf{CIFAR-10}} & \multicolumn{2}{c}{\textbf{Flowers-102}}& \multicolumn{2}{c}{\textbf{Stanford Cars}}\\
\cmidrule(l){2-3}
\cmidrule(l){4-5}
\cmidrule(l){6-7}
\cmidrule(l){8-9}
\cmidrule(l){10-11}
& Top1 & Top5& Top1& Top5& Top1 & Top5 & Top1 & Top5& Top1 & Top5\\
\midrule
AlignCLIP (TT)& 31.1 & 58.6 & 31.5 &60.9&64.8&95.5&18.7&39.5&10&36.4\\
\midrule
AlignCLIP (II)& \textbf{32.8} & \textbf{60.6} &  \textbf{36.5} & \textbf{66.4} & \textbf{69.3} & \textbf{97.8} & \textbf{18.8}& \textbf{40.3} &\textbf{11.8}&\textbf{38.1}\\
\midrule
AlignCLIP (II-TT) & 32.4 & 60 & 31.2 & 61.8 &66.4& 96.9 &18.7&39.9& 11.8 & 37.1 \\
\bottomrule
\end{tabular}
}
\end{sc}
\end{small}
\end{center}
\caption{Accuracy scores for zero-shot image classification.}
\label{res-table-zeroshot-cls-tt}
\end{table}

\begin{table}[t]
\vskip 0.15in
\begin{center}
\begin{small}
\begin{sc}
\resizebox{\linewidth}{!}{
\begin{tabular}{lcccccc}
\toprule
\textbf{Model} & \textbf{ImageNet-1K} &\textbf{CIFAR-100} & \textbf{CIFAR-10} & \textbf{Flowers-102}  & \textbf{Stanford Cars}\\
\midrule
AlignCLIP (TT)& 51.2 & 64.5 & 86.3 & 74.3 & 41.6\\
\midrule
AlignCLIP (II)& \textbf{51.5} & \textbf{67.4} & \textbf{87.2} & \textbf{76.8} & \textbf{45.6}\\
\midrule
AlignCLIP (II-TT) & 50.3 &  63.4  &  85.8 & 72.2& 42.2 \\
\bottomrule
\end{tabular}
}
\end{sc}
\end{small}
\end{center}
\caption{Accuracy scores for image classification with linear probing.}
\label{res-table-linear-probing-tt}
\end{table}

\subsection{Effects of text embedding separation}
\label{appendix-text-separation}
In Section \ref{method-imsep}, we proposed \imsep to enforce intra-modality separation among the image embeddings. A potential question that arises is that what happens if we enforce the separation amongst the text embeddings, or even, what happens of we enforce the separation inside the image embeddings as well as the text embeddings. To answer these questions, we trained versions of \myclip without image-image separation and with only text-text separation, namely \myclip-TT. In this case, we also used the re-scaling mechanism described in Section \ref{method-imsep}. Furthermore, we trained a version including both image-image embeddings separation and text-text embeddings separation, i.e., \myclip-II-TT. We followed the same experimental setup described in Section \ref{appendix-setup} and compare the results for zero-shot image classification as well as linear probing in Table~\ref{res-table-zeroshot-cls-tt} and Table~\ref{res-table-linear-probing-tt}, respectively. 

\section{Limitations}
\label{sec-limitations}
For the zero-shot classification experiments, we have tested the models using the ImageNet-1k, CIFAR-10, CIFAR-100, Stanford Cars, and Flowers-102 datasets. More classification analysis using other benchmarks datasets such as Food-100 and Hateful Memes has not been studied in this work. Furthermore, this study is limited to the English language and further analyses on multilingual and cross-lingual representation learning are necessary. Moreover, the re-scaling mechanism in the intra-modality separation loss is dependant on the choice of the pre-trained sentence encoder and has not been fully benchmarked in this work.
\section{Impact Statement}
\label{sec-impacts}
This paper presents work whose goal is to advance the field of Machine Learning at a fundamental level with regard to cross-modal representation learning. We acknowledge that this line of work has a broad range of potential societal implications. For instance, vision--language models may exhibit harmful biases and stereotypes, particularly when trained on data crawled from the Web. Due to their incorporation in prominent generative AI models, vision--language models may also contribute toward the model's ability to produce images portraying trademarked characters or notable figures. These sorts of concerns need to be carefully considered before incorporating such models into real-world applications.

\end{document}